\documentclass[conference]{IEEEtran}
\IEEEoverridecommandlockouts

\usepackage{CJKutf8}
\usepackage{setspace}
\usepackage{indentfirst}
\usepackage{float}
\usepackage{caption}
\usepackage{enumitem}
\usepackage{ulem}
\usepackage{array}
\usepackage{booktabs}
\usepackage{url}
\usepackage{subcaption}
\usepackage{booktabs}
\usepackage{amssymb}
\usepackage{multicol}
\usepackage{cite}
\usepackage{amsmath,amssymb,amsfonts}
\usepackage[linesnumbered,ruled,vlined]{algorithm2e}
\usepackage{algorithmic}
\usepackage{graphicx}
\usepackage{textcomp}
\usepackage{xcolor}
\usepackage{listings}
\usepackage{marvosym}

\def\BibTeX{{\rm B\kern-.05em{\sc i\kern-.025em b}\kern-.08em
    T\kern-.1667em\lower.7ex\hbox{E}\kern-.125emX}}

\DeclareRobustCommand*{\IEEEauthorrefmark}[1]{%
    \raisebox{0pt}[0pt][0pt]{\textsuperscript{\footnotesize\ensuremath{#1}}}}

\begin{document}

\title{Adaptive AI Agent Placement and Migration in Edge Intelligence Systems}

\author{
\IEEEauthorblockN{
Xingdan Wang\IEEEauthorrefmark{2,1},
Jiayi He\IEEEauthorrefmark{2,1},
Zhiqing Tang\IEEEauthorrefmark{1}\textsuperscript{\Letter},
Jianxiong Guo\IEEEauthorrefmark{1},\\
Jiong Lou\IEEEauthorrefmark{3},
Liping Qian\IEEEauthorrefmark{4},
Tian Wang\IEEEauthorrefmark{1},
Weijia Jia\IEEEauthorrefmark{1}
}
\IEEEauthorblockA{\IEEEauthorrefmark{1}Institute of Artificial Intelligence and Future Networks, Beijing Normal University, China}
\IEEEauthorblockA{\IEEEauthorrefmark{2}Faculty of Arts and Sciences, Beijing Normal University, China}
\IEEEauthorblockA{\IEEEauthorrefmark{3}Department of Computer Science and Engineering, Shanghai Jiao Tong University, China}
\IEEEauthorblockA{\IEEEauthorrefmark{4}College of Information Engineering, Zhejiang University of Technology, China}
\IEEEauthorblockA{\{wangxingdan, jiayihe\}@mail.bnu.edu.cn, \{jianxiongguo, zhiqingtang\}@bnu.edu.cn, lj1994@sjtu.edu.cn, \\lpqian@zjut.edu.cn, \{tianwang, jiawj\}@bnu.edu.cn}
\thanks{This work is supported by the National Natural Science Foundation of China (NSFC) under Grant 62302048. \textit{(Corresponding author: Zhiqing Tang.)}}
}

\maketitle

\begin{abstract}
The rise of LLMs such as ChatGPT and Claude fuels the need for AI agents capable of real-time task handling. However, migrating data-intensive, multi-modal edge workloads to cloud data centers, traditionally used for agent deployment, introduces significant latency. Deploying AI agents at the edge improves efficiency and reduces latency. However, edge environments present challenges due to limited and heterogeneous resources. Maintaining QoS for mobile users necessitates agent migration, which is complicated by the complexity of AI agents coordinating LLMs, task planning, memory, and external tools. This paper presents the first systematic deployment and management solution for LLM-based AI agents in dynamic edge environments. We propose a novel adaptive framework for AI agent placement and migration in edge intelligence systems. Our approach models resource constraints and latency/cost, leveraging ant colony algorithms and LLM-based optimization for efficient decision-making. It autonomously places agents to optimize resource utilization and QoS and enables lightweight agent migration by transferring only essential state. Implemented on a distributed system using AgentScope and validated across globally distributed edge servers, our solution significantly reduces deployment latency and migration costs. 
\end{abstract}


\section{Introduction}

The advancement of large language models (LLMs) and AI, such as ChatGPT, Claude, and DeepSeek, has made AI agents more convenient and effective at completing tasks, leading to their growing popularity \cite{deepseekai2025deepseekv3technicalreport}. Current research extensively explores AI agent deployment in cloud data centers to address various challenges, such as self-driving cars, intelligent robots, and voice assistants \cite{gao2024agentscopeflexiblerobustmultiagent}. In multi-modal edge scenarios, transmitting large volumes of edge user data, such as real-time video, point clouds, and sensor data, to the cloud introduces significant latency. Therefore, processing and aggregating data and tasks at the edge is very urgent \cite{8736011}. 

Deploying AI agents at the edge facilitates autonomous AI task planning and execution \cite{10970093}. This edge-based architecture offers scalability and supports greater device and user access compared to cloud-based deployments \cite{10217163}, offering many advantages. Firstly, it can reduce the distance of data transmission, thereby reducing latency and providing more real-time responses \cite{greengard2021internet}. Secondly, by planning and executing tasks locally, edge servers can reduce the amount of data transmitted to the cloud, thereby saving bandwidth and reducing costs \cite{9133107}. Furthermore, AI agents performing tasks on edge servers can appropriately reduce the frequency of sensitive data transmitted to the cloud, thereby improving data privacy and security \cite{WANG202075}. Finally, edge AI agents ensure continuous application operation and improves system reliability, even with unstable network connections \cite{tang2023multi}.

However, few studies address the deployment of AI agents in edge intelligence systems. Moreover, unlike the centralized processing in cloud data centers, the mobility of edge users necessitates the migration of AI agents to maintain optimal QoS, as agents should follow users as they move \cite{tang2018migration, tang2023multi}. Unlike traditional service deployment, AI agents often have various procedures such as LLM invocation, task planning, memory storage, and tool calling \cite{gao2024agentscopeflexiblerobustmultiagent}. Therefore, the placement and migration procedures of AI agents become much more complex \cite{mou2024adaptive,lou2022energy}. Traditional service or container deployment provides static services. Agent deployment, conversely, adaptively deploys agents, incorporating service caching, resource allocation, etc. \cite{8861020}. Deploying AI agents in edge intelligence systems is challenging due to complex environmental interactions and dynamics \cite{8744265}.

The first challenge is how to optimize AI agent placement considering the edge servers' limited computing, storage, and communication capabilities \cite{8746691}. Computing resources affect the decision-making and computation speed of agents. Storage resources affect the type of agents. For example, multi-modal agents must be deployed on edge servers with relatively abundant storage resources so that they can handle a large amount of data \cite{TULI2023103648}. Communication resources can affect the communication delay among agents. Agents heavily reliant on LLMs should be located on servers near their dependencies, whereas agents transferring large files benefit from servers with ample storage and communication resources.

The second challenge is how to consider the mobility requirements of edge users and migrate the agents accordingly. Dynamic environment configuration is crucial because the environmental context directly influences LLM-based agent responses \cite{WANG202075}. Maintaining environmental context integrity during and after migration is therefore essential \cite{9810299}. Moreover, unlike traditional service or container migration, AI agent migration only requires transferring the memory and configuration files, rather than the entire code base. This unique characteristic should be leveraged to minimize migration costs.

To address these challenges, we propose a novel framework for adaptive AI agent placement and migration in edge intelligence systems. We first model the AI agent in edge computing, fully considering the latency and cost during placement and migration, including transmission latency, initialization latency, migration latency, and computational costs, storage costs, etc. Then, based on the ant colony algorithm, we propose AI agent placement and migration algorithms, and introduce an LLM-based autonomous optimization scheme to further improve the algorithms' performance. We have implemented a distributed edge intelligence system based on AgentScope and deployed AI agents on edge servers distributed around the world \cite{gao2024agentscopeflexiblerobustmultiagent}. We implement the placement and migration of AI agents on the edge intelligence system, and the experimental results show that our algorithm reduces the deployment latency by 9.5\% and the migration cost by 11.5\% on average.

The main contributions are summarized as follows:

\begin{enumerate}
    \item We propose the deployment of AI agents in edge intelligence systems. Our objective is to minimize task execution and agent migration times while maximizing edge resource utilization.
    \item Our agent placement algorithm optimizes for computing and storage resources, and geographical distribution. Task scheduling and agent placement decisions are made based on the resource availability and current agent distribution.
    \item We design an adaptive AI agent migration algorithm, considering computing and communication resources. This algorithm autonomously decides whether to migrate the agent based on user movement, aiming to improve QoS.
    \item We have implemented our edge agent system and verified the effectiveness of our algorithms through a large number of experiments.
\end{enumerate}

\section{System Model and Problem Formulation}

\subsection{System Model}

\begin{figure}[t]
    \centering
    \includegraphics[width=0.75\linewidth]{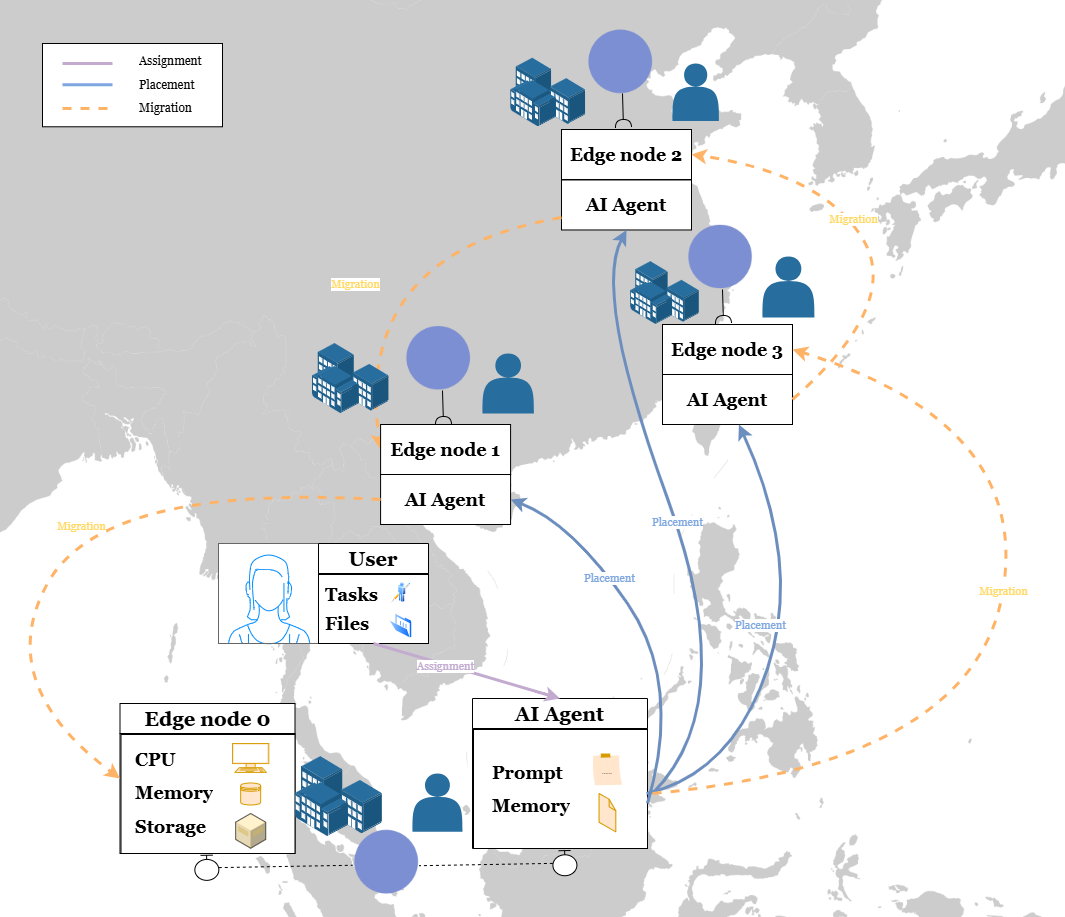}
    \caption{An overview of our edge intelligence system.}
    \label{fig:wide-fig}
\end{figure}

As shown in Fig. \ref{fig:wide-fig}, our edge intelligence system dynamically places and migrates AI agents to meet delay requirements of user tasks arriving at any time. Agents are deployed on edge servers for real-time processing. To minimize startup latency and communication latency due to user mobility, agents are strategically migrated. The core components are defined below.

\begin{enumerate}
    \item AI Agent: Dynamic and intelligent entities, denoted as $ a \in \mathbf{A} = \{a_1, a_2, \ldots, a_{|\mathbf{A}|}\}$, are deployed to different edge servers. Among them, $\left| \cdot \right|$ is used to indicate the number of elements in the set, and ${|\mathbf{A}|}$ is the number of AI agents. Each AI Agent $a$ has a memory requirement $M_a$ (GB), computing resource requirements $C_a$ (CPU cycles), storage resource requirements $S_a$ (GB), communication resource requirements $V_a$ (Mbps), edge server location $E_a$ (server number), etc.
    \item Edge System: The edge server set is defined as $\mathbf{E} = \{e_1, e_2, \ldots, e_{|\mathbf{E}|}\}$. Each edge server $e \in \mathbf{E}$ has resource capacities such as memory capacity $M_e$ (GB), computing resource capacity $C_e$ (CPU cycles), storage resource capacity $S_e$ (GB), and communication resource capacity $V_e$ (Mbps). Furthermore, the AI Agent running on each edge server \(e \) is defined as the set $\mathbf{A}_e $.
    \item Task: The task currently being executed is defined as the set $R = \{r_1, r_2, \ldots, r_{|\mathbf{R}|}\}$. Each user task $r$ has a prompt $p_r$ and file $f_r$, and requires a set of AI Agents $\mathbf{A}_r$.
\end{enumerate}


\subsection{Cost}

\subsubsection{Latency}

The latency encompasses file transfer, agent startup, migration, task processing, and result return.

\textbf{Transmit Latency:} It comprises file upload/reception and prompt transmission, defined as:
\begin{equation}
\begin{matrix}
T_r^{\text{tra}} = \frac{s_r}{\eta_e} + T_p,
\end{matrix}
\end{equation}
where $s_r$ represents the size of the file, $\eta_e$ represents the actual network bandwidth of the edge node, and $T_p$ represents the time of receiving the prompt.

\textbf{Migration Latency:} Migration involves exporting and transferring memory, followed by initiating the AI agent and importing the memory. Memory transfer time is calculated as:
\begin{equation}
\begin{matrix}
T_m^{tra} = \frac{M_a}{\eta_{e_1}^{e_2}},
\end{matrix}
\end{equation}
where $M_a$ represents memory size and $\eta_{e_1}^{e_2}$ denotes the network bandwidth during memory transmission. The migration delay is defined as:
\begin{equation}
T_r^{\text{mig}} = T_{e_1}^\text{export} + T_m^{tra} + T_{e_2}^\text{initiation} +  T_{e_2}^\text{load},
\end{equation}
where $T_{e_1}^\text{export}$ is the time when the edge node $e_1$ exports memory, and $T_{e_2}^\text{initiation}$ is the time when the edge node $e_2$ initiates the AI agent. $T_{e_2}^\text{load}$ is the time when the edge node $e_2$ is imported into memory. 

\textbf{Initiation Latency:} The initial delay, caused by the aggregate startup times of all AI agents, increases with the number of agents. Therefore, it is defined as:
\begin{equation}
\begin{matrix}
T_r^{\text{Ini}} = \sum_{a \in \mathbf{A_r}} T_{\text{init}}\left(|\mathbf{A_{e(a)}}|\right),
\end{matrix}
\end{equation}
where $A_r$ is the set of AI agents required for task $r$, and $e_a$is the server where AI Agent a is located.

\textbf{Processing Latency:} It depends on task requirements. If file processing is necessary, latency varies based on file type and quantity. Otherwise, only LLM reasoning time is relevant:
\begin{equation}
T_r^{\text{pro}} = T_{fpro} + T_{LLM},
\end{equation}
where $T_{fpro}$ refers to the file processing time, and $T_{LLM}$ refers to the thinking time of LLM, indicating that total processing delay comprises both file processing and LLM computation. Finally, the total task duration is defined as:
\begin{equation}
T_r = T_r^{\text{tra}} + T_r^{\text{mig}} + T_r^{\text{Ini}} + T_r^{\text{pro}}.
\end{equation}

\subsubsection{Cost} 

The cost reflects resource utilization, encompassing computing, storage, and communication usage.

\textbf{Computing Cost:} Computing resources refer to CPU usage:
\begin{equation}
\begin{matrix}
O_r^{CPU} = T_r - (\frac{s_r}{\eta_e} + T_m^{\text{tra}}).
\end{matrix}
\end{equation}

\textbf{Storage Cost:} It encompasses all necessary task files and the AI agent's memory, defined as:
\begin{equation}
\begin{matrix}
O_r^{Stor} = \sum_{i=1}^{|\mathbf{R}|} S_r + \sum_{j=1}^{|\mathbf{A}|} M_a,
\end{matrix}
\end{equation}
where $|\mathbf{R}|$ is the number of tasks, $S_r$ is the storage space for task $r$, and $M_a$ is the memory usage of AI Agent $a$.

\textbf{Communication Cost:} Interdependent AI agents communicate and transfer files using communication resources:
\begin{equation}
\begin{matrix}
O_r^{Comm} = \sum_{i=1}^{m} C_i,
\end{matrix}
\end{equation}
where $C_i$ represents the communication resource usage for each of the $m$ events. The total cost is defined as:
\begin{equation}
O_r = O_r^{CPU} + O_r^{Stor} + O_r^{Comm}.
\end{equation}

\subsubsection{Constraints} For each agent $a$ assigned to edge server $e$, resource requirements must be met:
\begin{align}
\begin{matrix}
    \quad \sum_{a \in \mathbf{A}_r} M_a z_{a,e} \leq M_e^{\text{rem}}, \quad \forall e \in \mathcal{N}(e_c), \quad \\
    \quad \sum_{a \in \mathbf{A}_r} S_a z_{a,e} \leq S_e^{\text{rem}}, \quad \forall e \in \mathcal{N}(e_c), \quad \\
    \quad \sum_{a \in \mathbf{A}_r} V_a z_{a,e} \leq V_e^{\text{rem}}, \quad \forall e \in \mathcal{N}(e_c), \quad
\end{matrix}
\end{align}

\subsubsection{Problem Formulation} 

The Edge Agent Deployment (EAD) problem is defined as:

\textbf{Problem EAD.}
\begin{align*}
\begin{matrix}
    \min_{z} & \quad \sum_{a \in \mathbf{A}_r} \left( \frac{M_a}{\eta_{e_c}^e} + T_e^{\text{init}} \right) + \sum_{e \in \mathcal{N}(e_c)} \left( \frac{\sum_a C_a z_{a,e}}{C_e^{\text{rem}}} \right) \\
    & \quad + \theta \sum_{a \in \mathbf{A}_r} \sum_{(a_i,a_j)\in \mathcal{D}_a} T_{\text{comm}}(e_{a_i}, e_{a_j}) \\
    \text{s.t.} 
    & \quad \sum_{e \in \mathcal{N}(e_c)} z_{a,e} = 1, \quad \forall a \in \mathbf{A}_r \quad \\
    & \quad z_{a_i,e} = z_{a_j,e}, \quad \forall (a_i,a_j) \in \mathcal{D}_a, \forall a \in \mathbf{A}_r \\
    & \quad
\end{matrix}
\end{align*}
The EAD problem seeks to minimize initial expected cost, considering resource constraints. Moreover, the Edge Agent Migration (EAM) problem is:

\textbf{Problem EAM.}
\begin{align*}
\begin{matrix}
\text{Net } & \text{Gain} (e_a \rightarrow e') =  \underbrace{\Delta T_{\text{latency}}}_{\text{delay benefits}} - \underbrace{(T_{\text{mig}} + \gamma O_{\text{overhead}})}_{\text{migration costs}} \\
& - \underbrace{\theta_{\text{dep}} \cdot \Delta T_{\text{dep}}}_{\text{dependency damage costs}} \\
\text{s.t.}
& \Delta T_{\text{latency}} = \; \mathbb{E}[T_{\text{comm}}(e_c, e_a)] - \mathbb{E}[T_{\text{comm}}(e_c, e')] \\
& \Delta T_{\text{dep}} = \; \sum_{a' \in \mathcal{D}_a} \left( T_{\text{comm}}(e', e_{a'}) - T_{\text{comm}}(e_a, e_{a'}) \right)
\end{matrix}
\end{align*}

The EAM problem is triggered when:
\begin{enumerate}
    \item \textbf{Position Deviation Threshold:} The number of hops $H(e_c, e_a)$ to the server $e_a$ hosting the AI Agent, caused by the client's movement, exceeds $H_{th}$.
    \item \textbf{Resource Bottleneck Warning:} The available resources of server $e_a$ fall below the security threshold.
\end{enumerate}

Our edge AI system, addressing both EAD and EAM, can be modeled as a time-scaled boxing problem. Initial deployment allocates resources, and subsequent migration dynamically adjusts AI agent placement within edge server capacity limits.

\section{Algorithms}

We present AntLLM, an algorithm comprising AntLLM Placement (ALP) and AntLLM Migration (ALM), which uses the Ant Colony Algorithm for initial placement and migration, then refines these decisions with LLM assistance for optimized placement and migration strategies.

\subsection{AntLLM Placement Algorithm (ALP)}

\begin{algorithm}[t]
\SetAlgoLined
\KwIn{Task list $T$, Edge server set $E$, Resource status $R$, LLM model $M_{LLM}$}
\KwOut{Final deployment strategy $S_{\text{final}}$}

Initialize pheromone matrix $\tau$ and heuristic matrix $\eta$\;

\For{each ant $k = 1$ to $m$}{
    \For{each task $t_i \in T$}{
        Select a server $e_j \in E'$ using probability:
        $$
        p_{ij} = \frac{[\tau_{ij}]^\alpha \cdot [\eta_{ij}]^\beta}{\sum_{e \in E'} [\tau_{ie}]^\alpha \cdot [\eta_{ie}]^\beta}
        $$
        Assign task $t_i$ to server $e_j$\;
    }
    Compute deployment cost $C_k$\;
}
Select the best deployment: $S^* = \arg\min_k C_k$\;
Update pheromone matrix $\tau$\;

\textbf{Post-Optimization:} Use $M_{LLM}$ to refine and validate $S^*$\;
Obtain final optimized deployment $S_{\text{final}}$\;

\caption{ALP}\label{algorithm_alp}
\end{algorithm}

The ALP algorithm's process is shown in Algorithm \ref{algorithm_alp}, and the main process is as follows.

\begin{enumerate}
    \item Path Modeling. Frame deployment as a path selection problem: ants navigate from the first agent, choosing an edge server for each agent. Path nodes are represented as $(a, e)$, and path length is ${| \mathbf {A_r} |} $.
    \item Pheromone Matrix. A two-dimensional matrix, $pheromone[a][e]$, represents the attractiveness of assigning AI agent $a$ to edge node $e$.
    \item Heuristic Function. The heuristic function $\eta$, based on Problem EAD, evaluates a server's suitability for an agent. It combines an initial score, a resource score, and a communication score.
    \item Probability Selection Rules. The probability of an ant agent selecting deployment target $e$ is determined by:
\begin{equation}
\begin{matrix}
P_{a,e} = \frac{[\tau_{a,e}]^\alpha \cdot [\eta_{a,e}]^\beta}{\sum_{e'} [\tau_{a,e'}]^\alpha \cdot [\eta_{a,e'}]^\beta},
\end{matrix}
\end{equation}
where $\tau_{a,e}$ represents the pheromone concentration deposited by agent $a$ at node $e$, and $\eta_{a,e}$ is a heuristic value. Parameters $\alpha$ and $\beta$ weigh the influence of pheromones and the heuristic, respectively.
    \item Path Construction. Each ant builds a complete deployment plan, verifying resource limits and avoiding illegal plans during construction.
    \item Path Evaluation. Score each deployment path based on: total deployment delay, total resource expenditure, and communication dependency cost between agents.
    \item Pheromone Update. After all ants have completed deployment, update the pheromone matrix. Pheromones volatilize, and trails are reinforced based on evaluation:
\begin{equation}
\begin{matrix}
\tau_{a,e} \leftarrow (1-\rho) \cdot \tau_{a,e} + \Delta \tau_{a,e},
\end{matrix}
\end{equation}
    \item Iterative Update. Iterate pheromone construction, evaluation, and update to determine the optimal placement.
\end{enumerate}

\subsection{AntLLM Migration Algorithm (ALM)}

\begin{algorithm}[t]
\SetAlgoLined
\KwIn{Current deployment $S$, Resource status $R$, Performance threshold $C$, LLM model $M_{LLM}$}
\KwOut{Final migration plan $M_{\text{final}}$}

\If{performance degradation condition $C$ is triggered}{

    Initialize pheromone matrix $\tau$ and heuristic matrix $\eta$\;

    \For{each ant $k = 1$ to $m$}{
        Initialize empty migration plan $M_k$\;

        \For{each task $t_i$ to migrate}{
            Select a target server $e_j \in E_{cand}$\;
            Assign $t_i$ to $e_j$, update $M_k$\;
        }
        Compute migration cost $C_k$\;
    }
    Select the optimal plan: $M^* = \arg\min_k C_k$\;
    Update pheromone matrix $\tau$\;

    \textbf{Post-Optimization:} Use $M_{LLM}$ to refine and validate $M^*$\;
    Obtain final optimized migration plan $M_{\text{final}}$\;
}

\caption{ALM}\label{algorithm_alm}
\end{algorithm}

The ALM algorithm is illustreted in Algorithm \ref{algorithm_alm}. The details are as follows.

\begin{enumerate}
    \item State Space Modeling. The search space consists of all transferable nodes $N(e_a)$ adjacent to the agent's current server $e_a$. Each ant attempts to migrate from $e_a$ to a server $e'$ within this space.
    \item Pheromone Initialization. Initialize the pheromone matrix $\tau_{e'}$ to reflect target node attractiveness.
    \item Heuristic Function. The heuristic $\eta$ uses Problem EAM to optimize latency reduction, migration time, and dependency latency.
    \item Probabilistic Transfer Selection. Each ant selects the target server according to the following equation:
\begin{equation}
\begin{matrix}
P_{e'} = \frac{[\tau_{e'}]^\alpha \cdot [\eta_{e'}]^\beta}{\sum_{e'' \in N(e_a)} [\tau_{e''}]^\alpha \cdot  [\eta_{e''}]^\beta},
\end{matrix}
\end{equation}
where $\alpha$ and $\beta$ represent the importance of pheromones and inspiring information, respectively.
    \item Construction of the Migration Plan. Each ant plans its migration. If the target node $e'$ lacks sufficient resources, the plan is invalid and penalized.
    \item NetGain Evaluation. Prioritize schemes with higher migration benefits:
\begin{equation}
\begin{matrix}
\text{Fitness}(e') = \Delta T_{\text{latency}} - (T_{\text{mig}} + \gamma O_{\text{overhead}}),
\end{matrix}
\end{equation}
Fitness decreases if resource constraints are violated.
    \item Pheromone Update. It evaporates after each ant traversal:
\begin{equation}
\begin{matrix}
\tau_{e'} \leftarrow (1-\rho) \cdot \tau_{e'},
\end{matrix}
\end{equation}
Obtain the optimal scheme reinforcement:
\begin{equation}
\begin{matrix}
\tau_{e'} \leftarrow \tau_{e'} + \frac{Q}{1 + \text{Cost}(e')},
\end{matrix}
\end{equation}
    \item Iterative optimization. Iterate the process and output the optimal migration target server, $e^*$.
\end{enumerate}

\section{System Implementation}

Our system is implemented through AgentScope \cite{gao2024agentscopeflexiblerobustmultiagent}. The following outlines the step-by-step system planning for deployment and migration, enabling autonomous deployment and migration of the distributed AI agent system.

\subsection{System Deployment Steps}

\begin{figure*}[t]
    \centering
    \begin{tabular}{cccc}
        \includegraphics[width=0.21\textwidth]{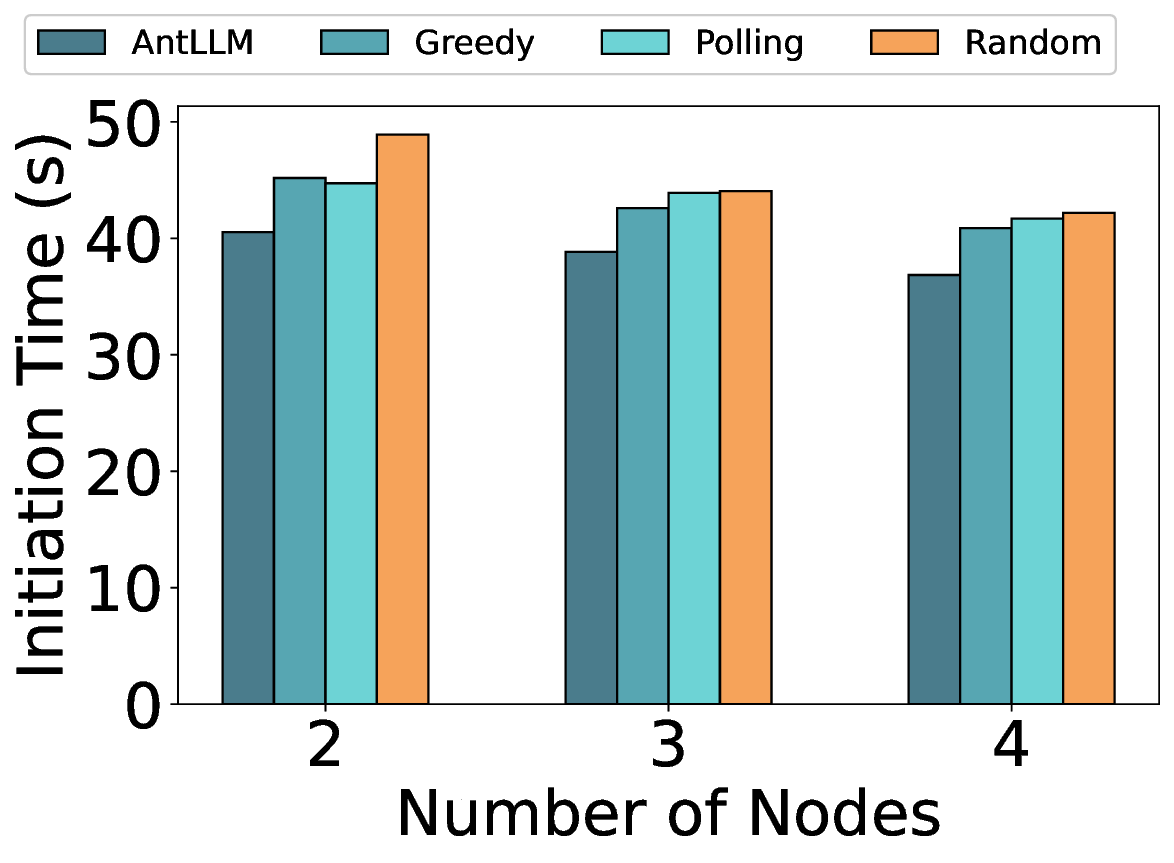} & 
        \includegraphics[width=0.21\textwidth]{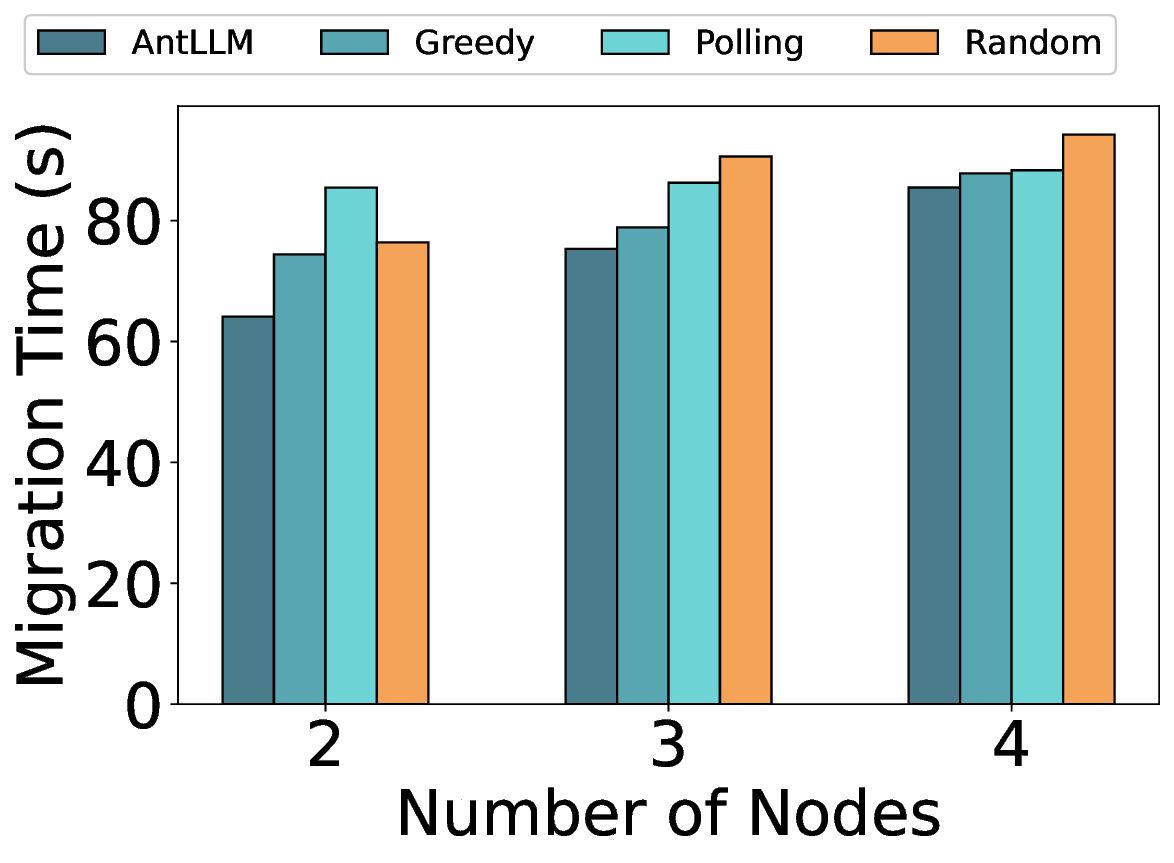} & 
        \includegraphics[width=0.21\textwidth]{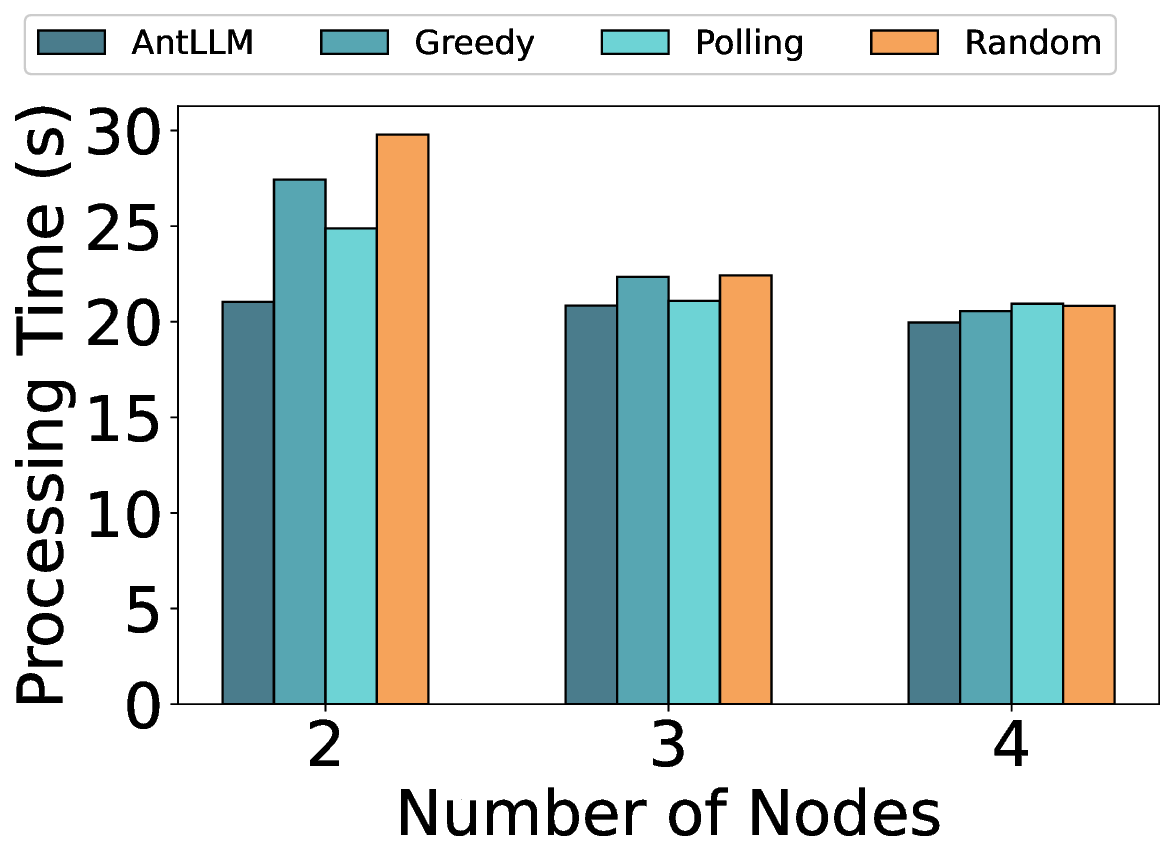} &
        \includegraphics[width=0.21\textwidth]{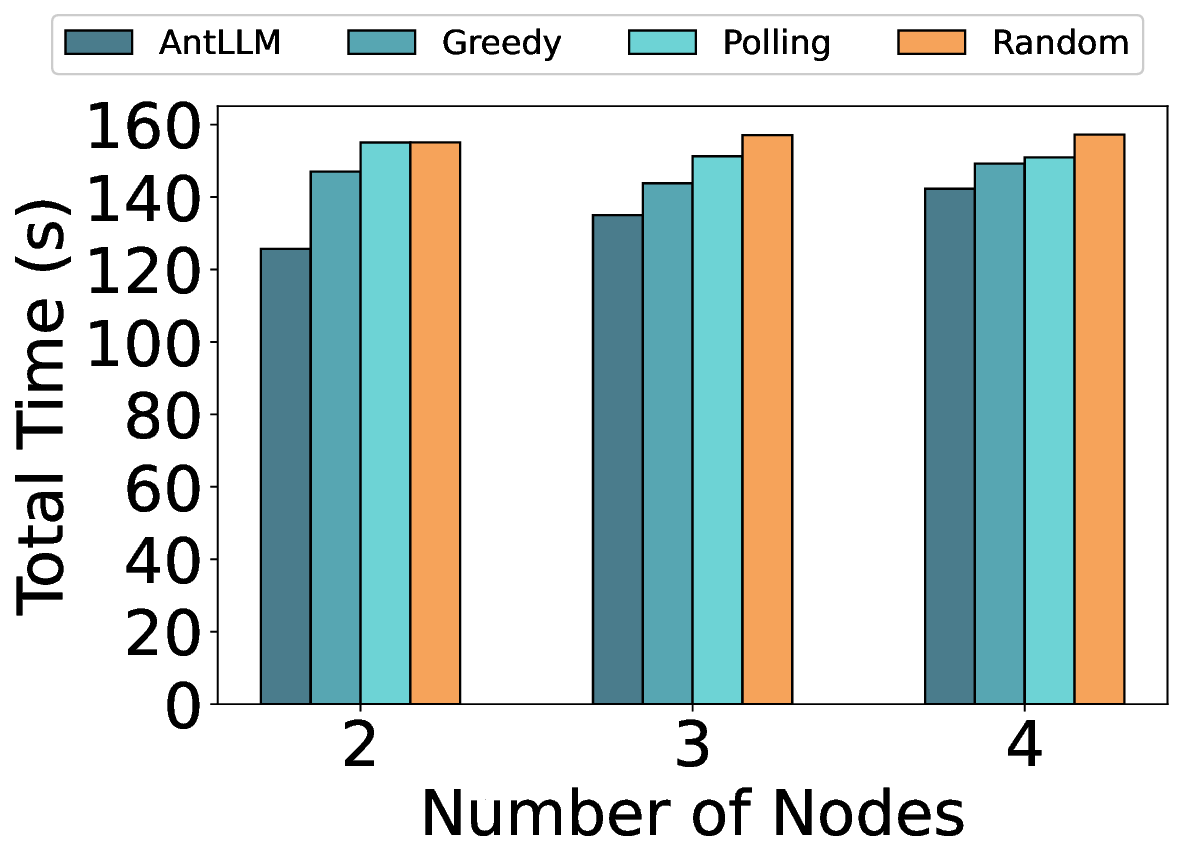} \\
        (a) Initial Time & (b) Migration Time & (c) Process Time & (d) Total Time
    \end{tabular}
    \caption{Delay Performance with different number of nodes}
    \label{fig:Delay Performance with different number of nodes}
\end{figure*}

\begin{figure*}[t]
    \centering
    \begin{tabular}{cccc}
        \includegraphics[width=0.21\textwidth]{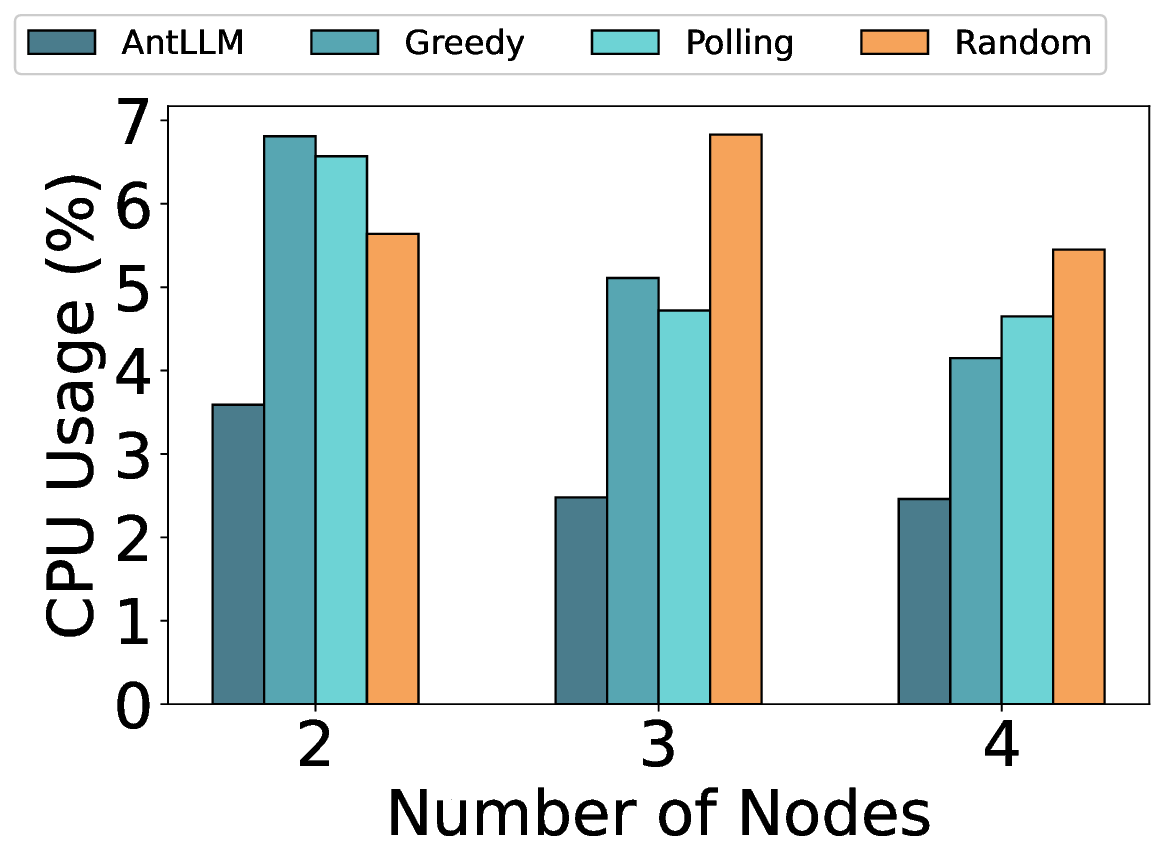} & 
        \includegraphics[width=0.21\textwidth]{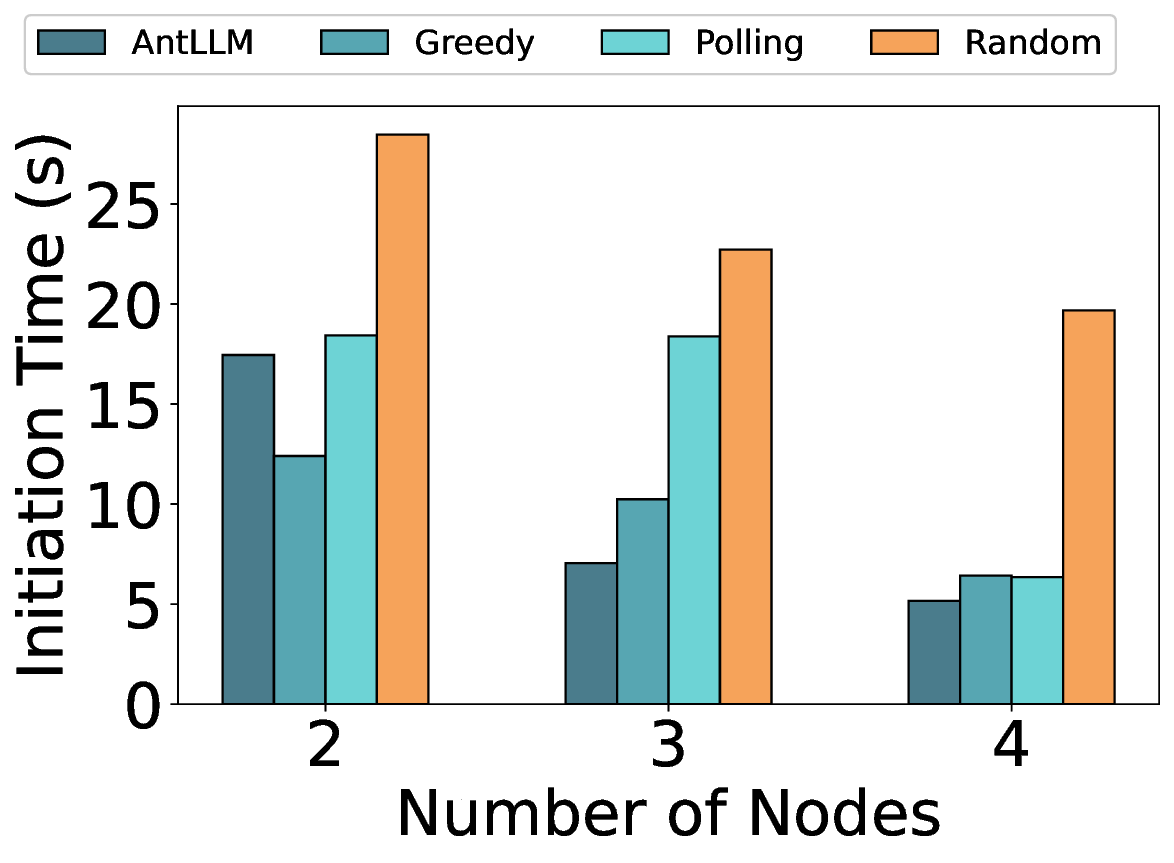} & 
        \includegraphics[width=0.21\textwidth]{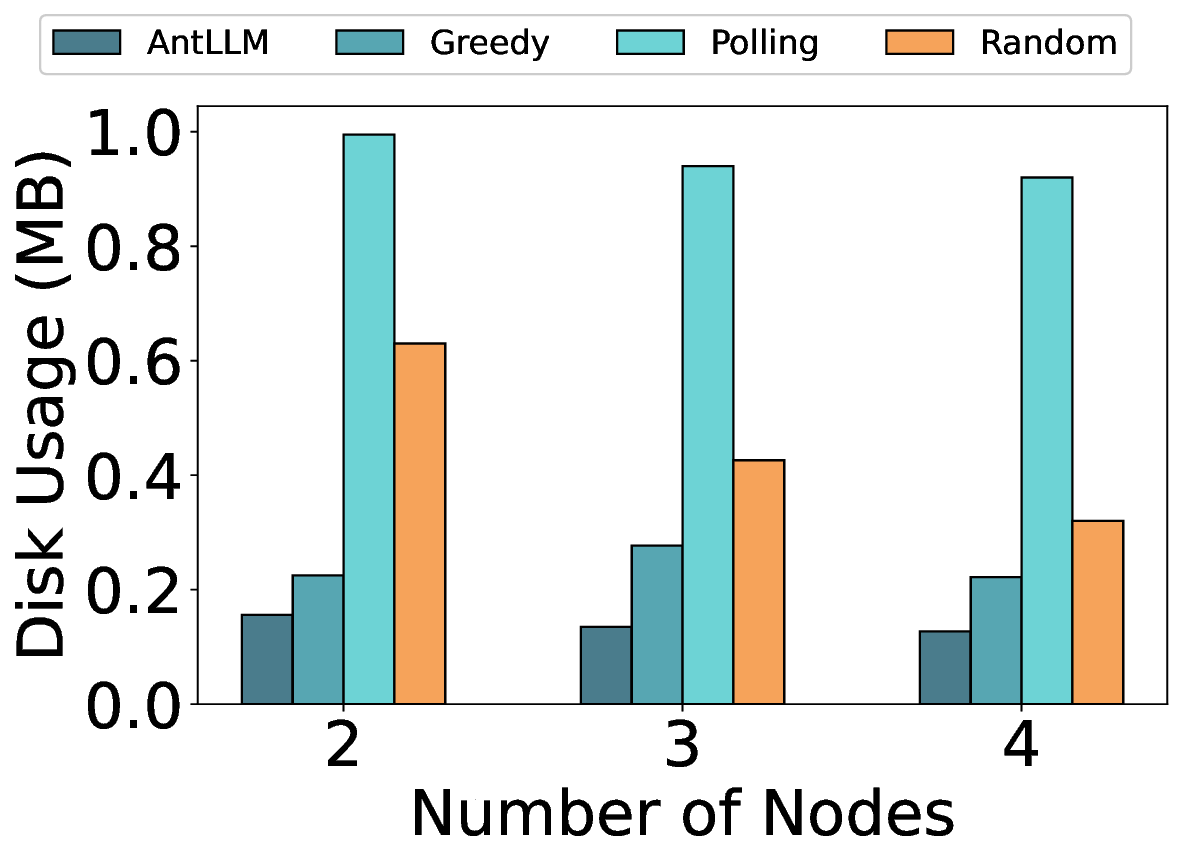} &
        \includegraphics[width=0.21\textwidth]{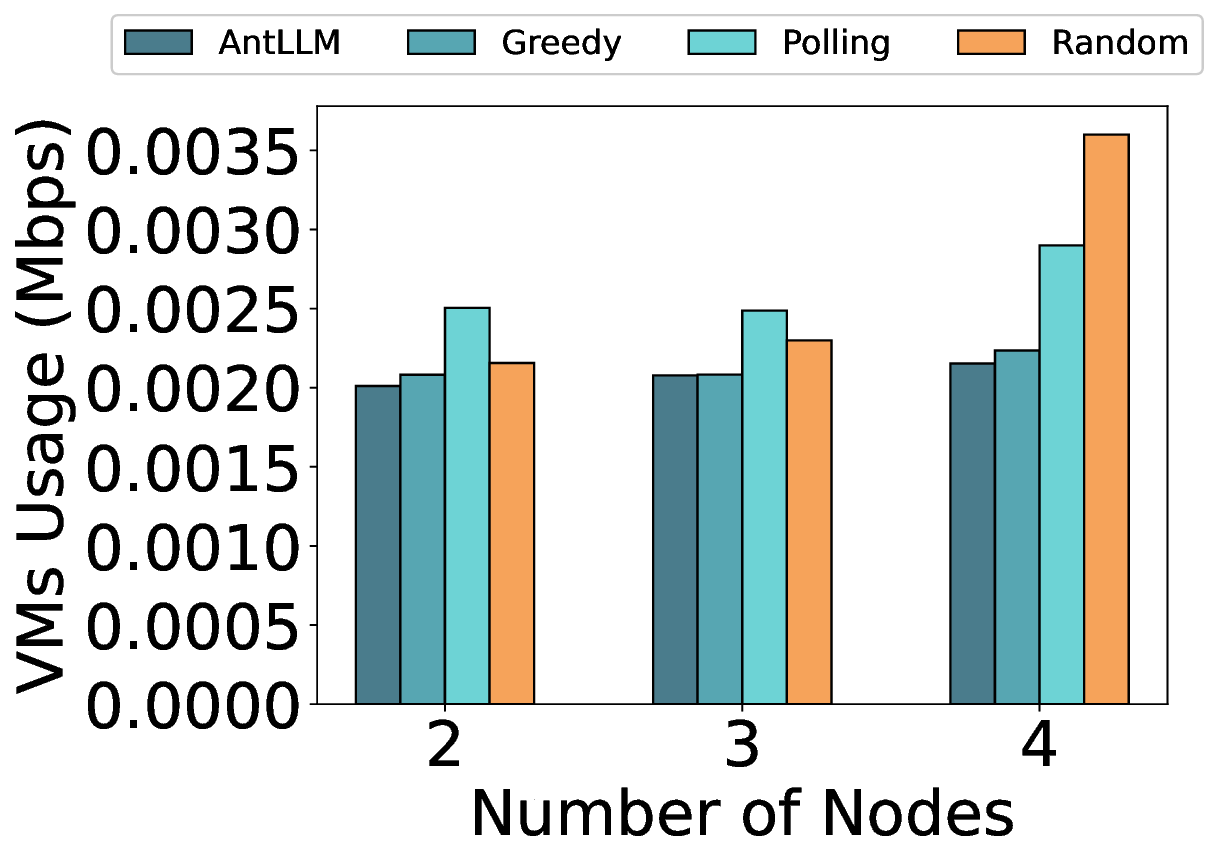} \\
        (a) CPU Usage & (b) Memory Usage & (c) Disk Usage & (d) VMs Usage
    \end{tabular}
    \caption{Cost Performance with different number of nodes}
    \label{fig:Overhead Performance with different number of nodes}
\end{figure*}

\begin{enumerate}
    \item Task Analysis and Planning. An AI agent analyzes the user's task document and uses a \verb|DialogAgent| to automatically refine the goal, decompose it into sub-tasks handled by multiple agents, generate resource requirements (CPU/Memory/Storage/Network) for each agent, and establish agent dependencies.
    \item Obtain Real-time Server Resources. \verb|Paramiko| is used to connect to the edge server and execute remote commands, retrieving real-time resources. Specifically, network throughput is calculated by reading \verb|/proc/net/dev| twice.
  
    \item Optimal Resource Allocation. We calculate the server efficiency score using a defined function, then optimize deployment solutions with the AntLLM algorithm to generate the best option.

    \item Remote Service Deployment. We start the AI agent by executing remote commands via SSH, verifying successful startup by checking the process ID.

    \item Task Execution Coordination. Recursively invoke the execute function to process dependencies.

\end{enumerate}

\subsection{Dynamic Migration Steps} 

The system resumes the task after dynamic migration.

\begin{enumerate}

    \item Triggering Conditions. Driven by user motivation, periodic detection, and event triggers.

    \item Migration Decision. Re-evaluate candidate servers using the fetch server function, then determine the migration decision using the AntLLM algorithm.

    \item Migration Execution. Migrate agent memory data via export/import functions to initiate a new instance on the target server while releasing the original resource.

\end{enumerate}

\section{Experiments}

\begin{figure*}[t]
    \centering
    \begin{tabular}{cccc}
        \includegraphics[width=0.21\textwidth]{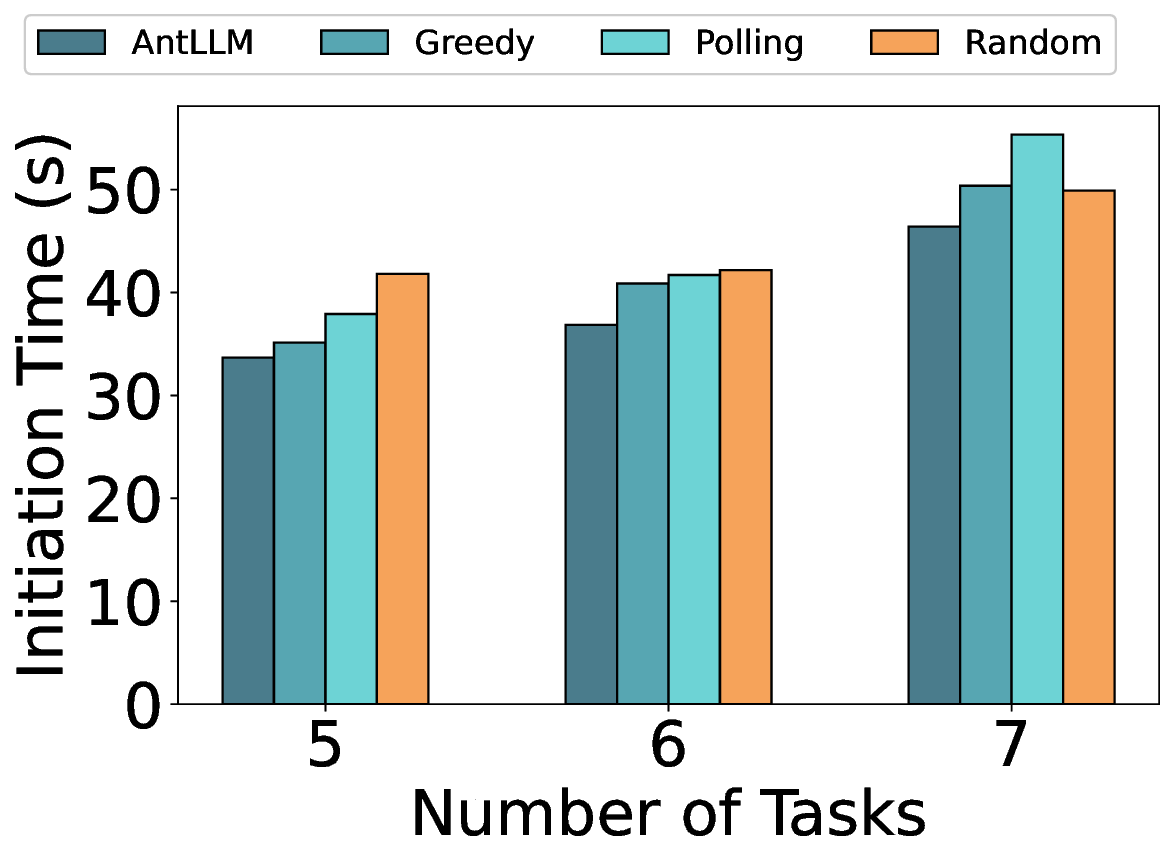} & 
        \includegraphics[width=0.21\textwidth]{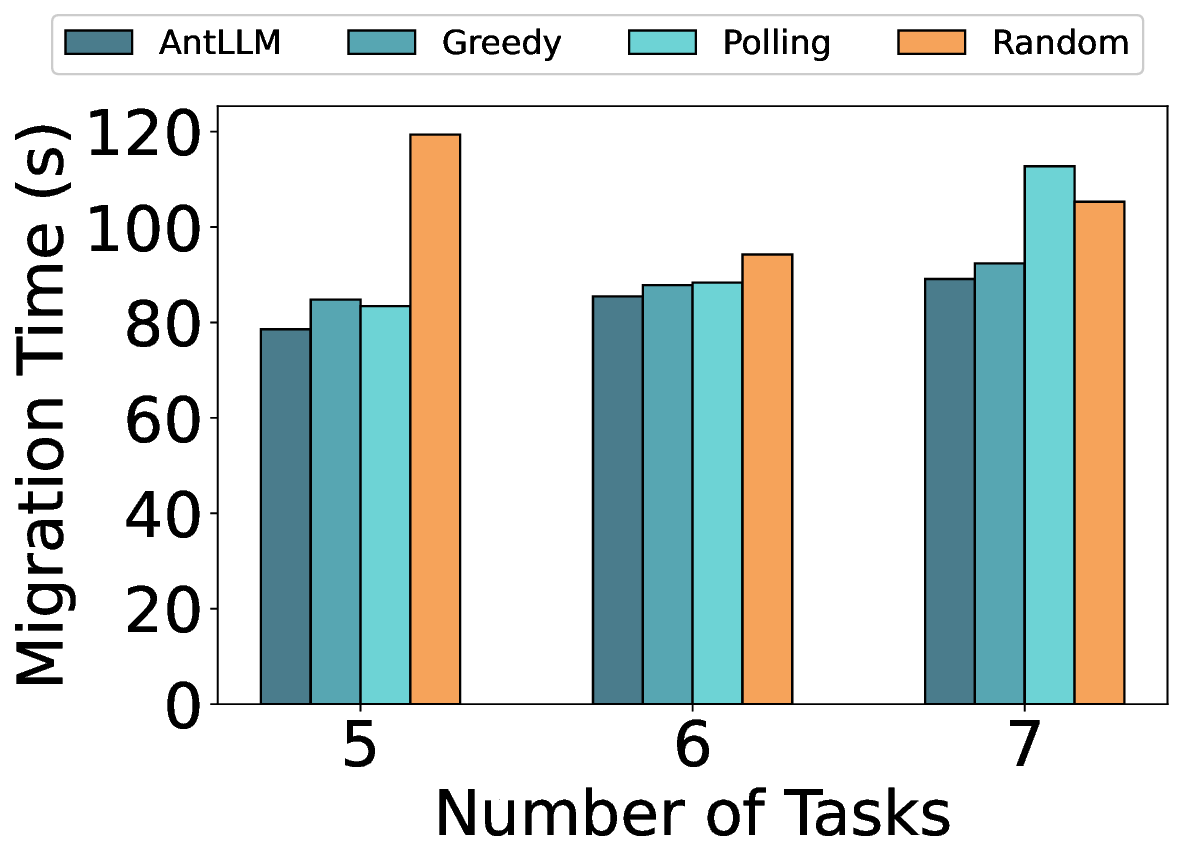} & 
        \includegraphics[width=0.21\textwidth]{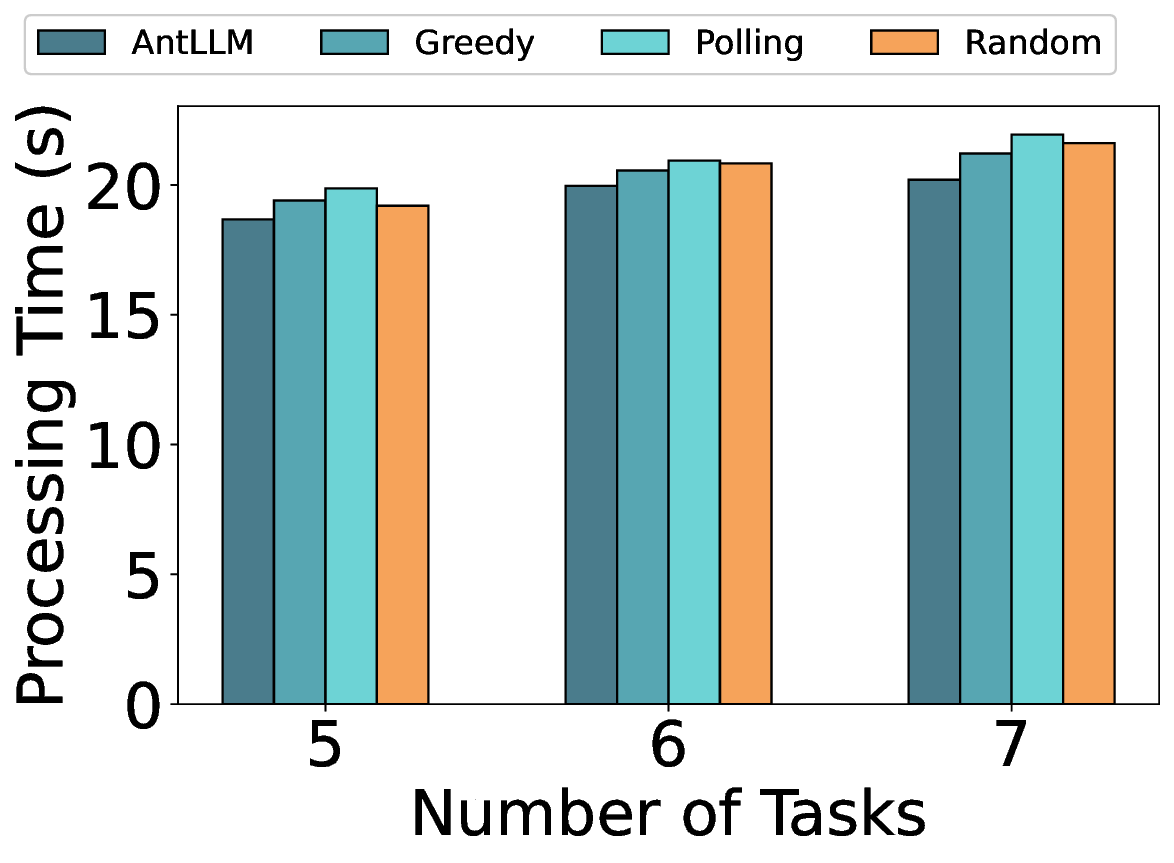} &
        \includegraphics[width=0.21\textwidth]{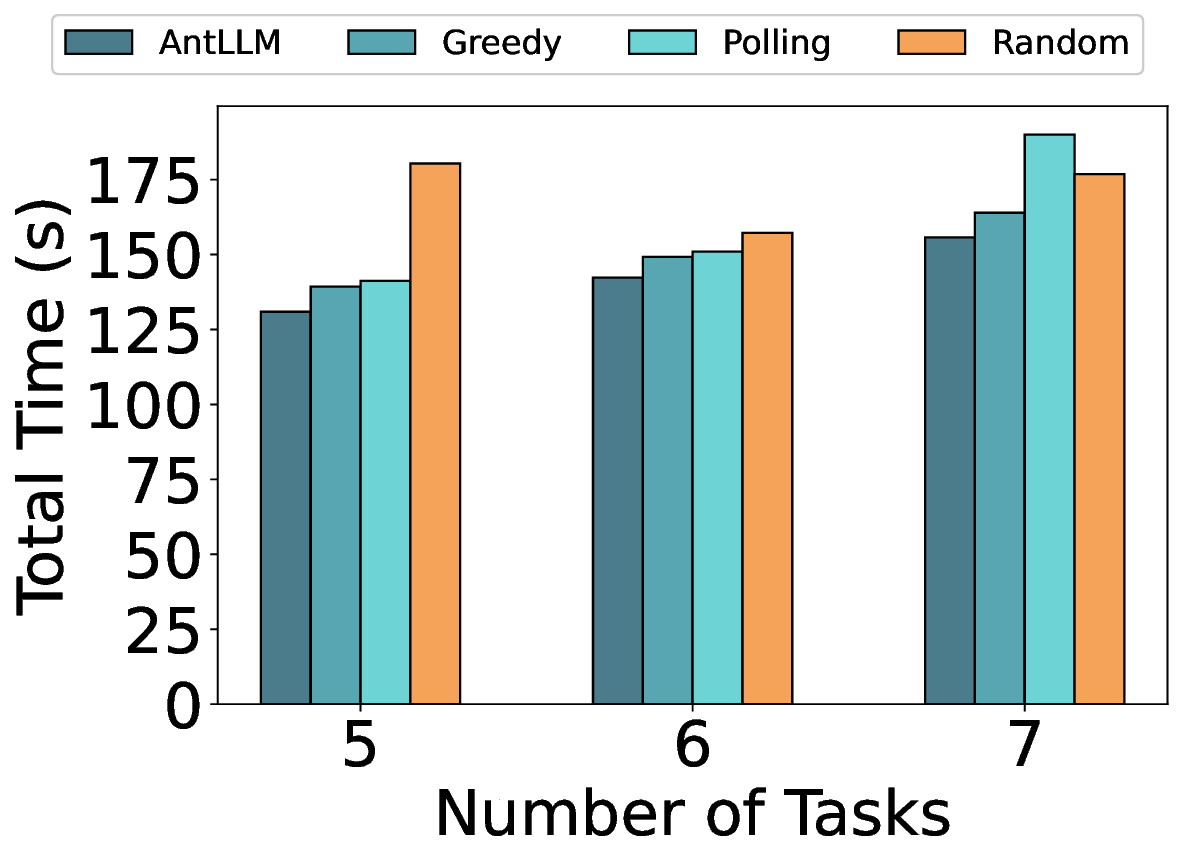} \\
        (a) Initial Time & (b) Migration Time & (c) Process Time & (d) Total Time
    \end{tabular}
    \caption{Delay Performance with different number of tasks}
    \label{fig:Delay Performance with different number of tasks}
\end{figure*}

\begin{figure*}[t]
    \centering
    \begin{tabular}{cccc}
        \includegraphics[width=0.21\textwidth]{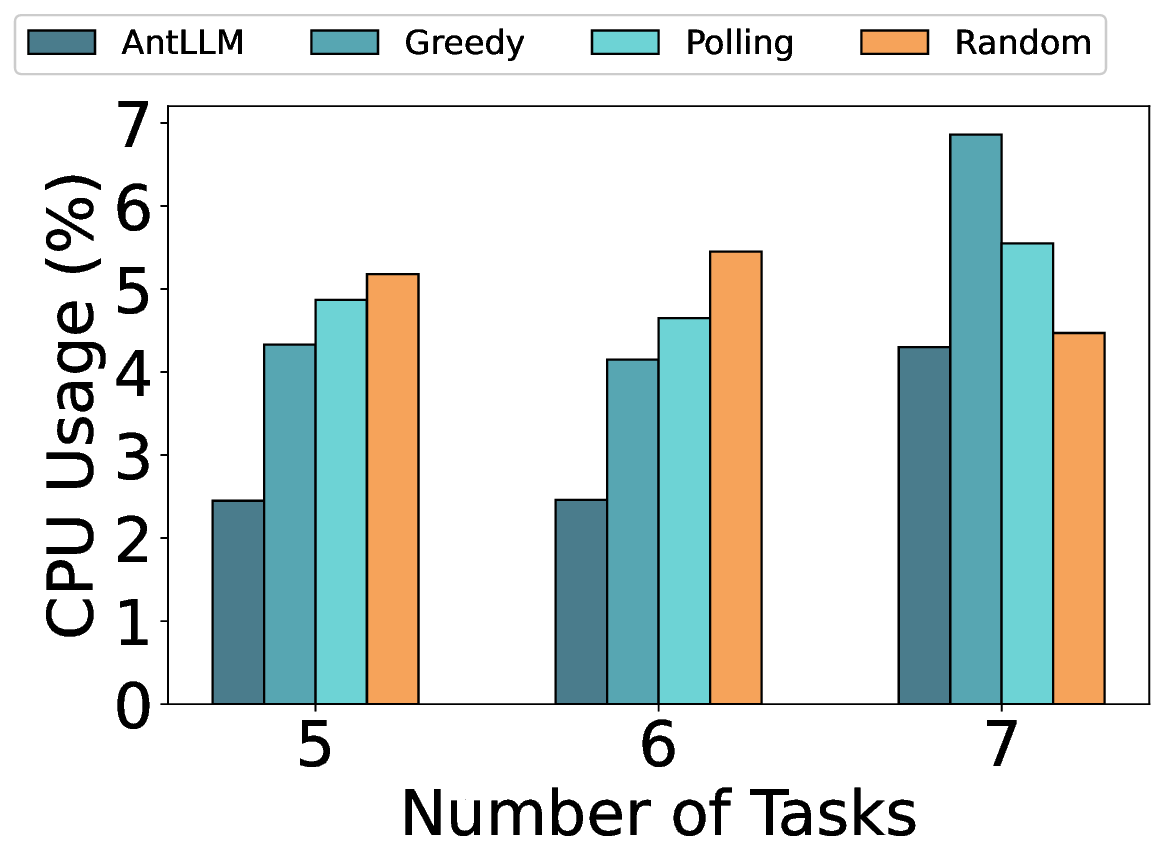} & 
        \includegraphics[width=0.21\textwidth]{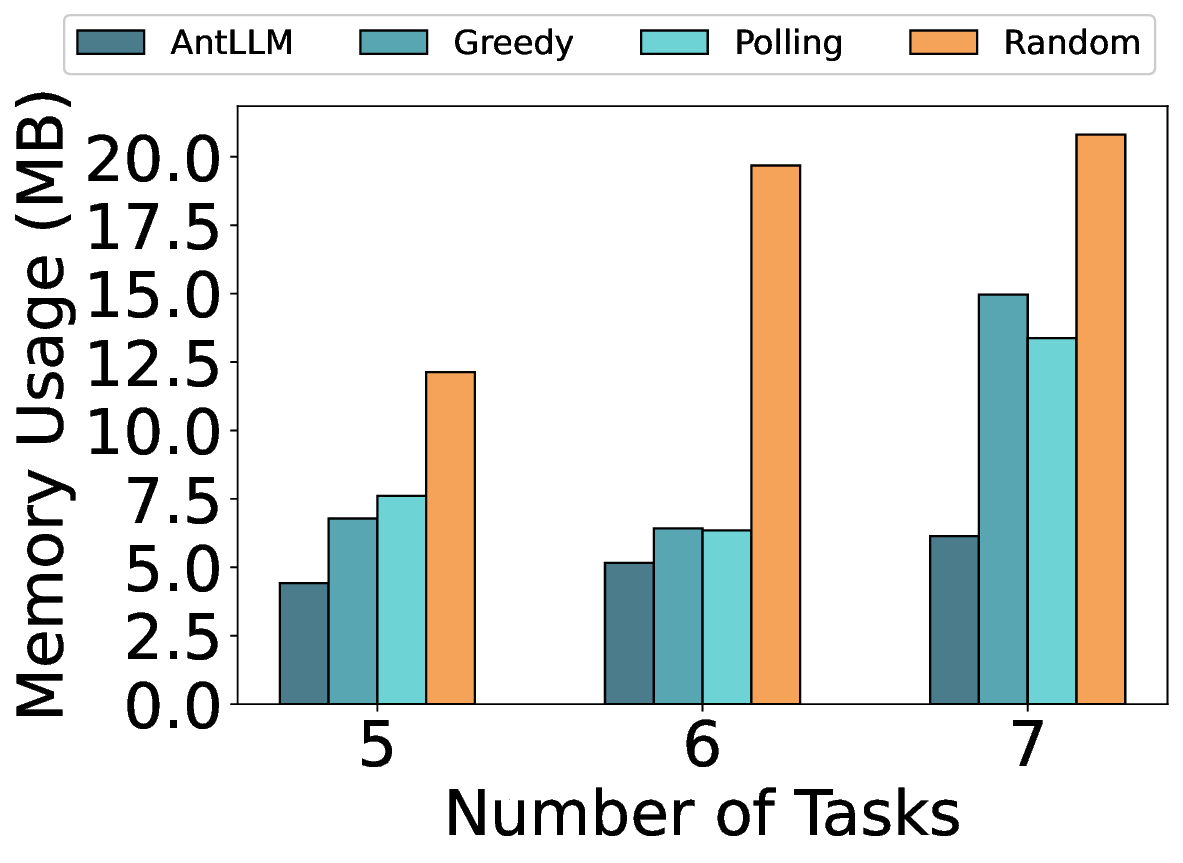} & 
        \includegraphics[width=0.21\textwidth]{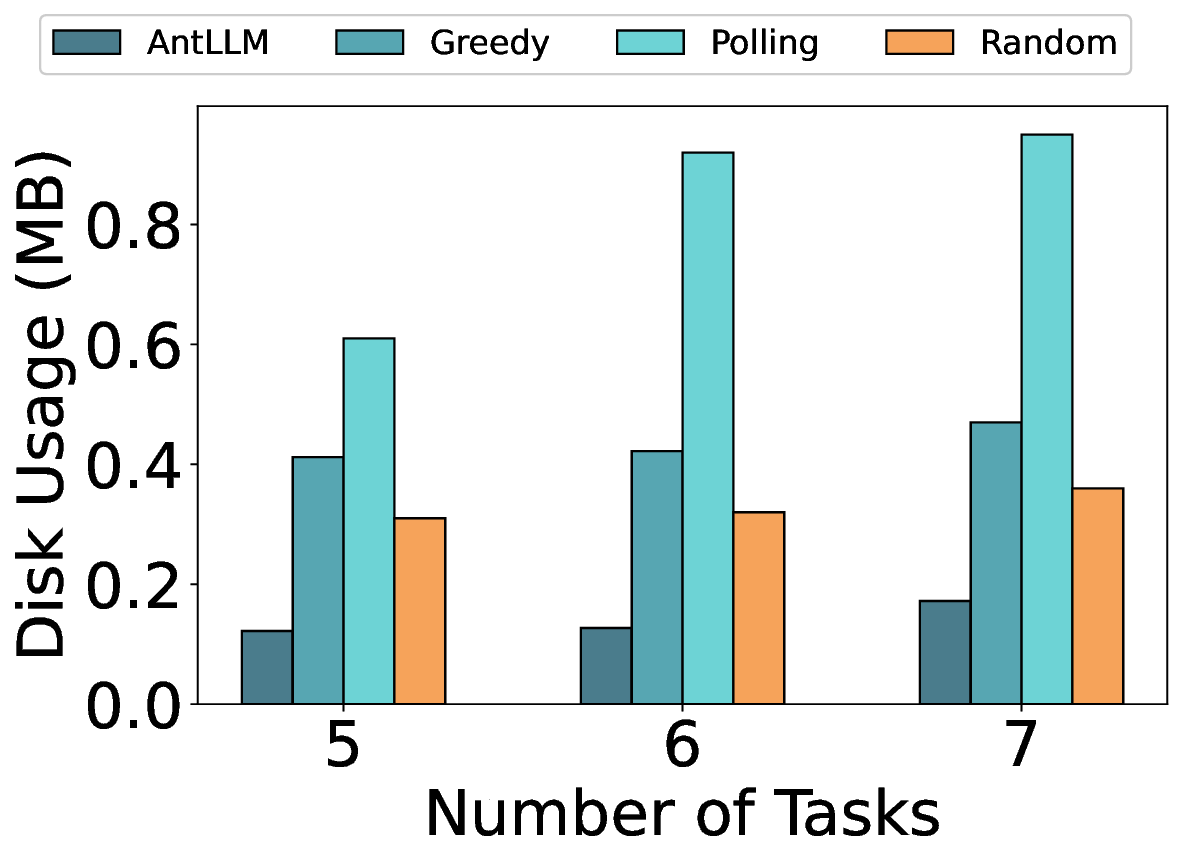} &
        \includegraphics[width=0.21\textwidth]{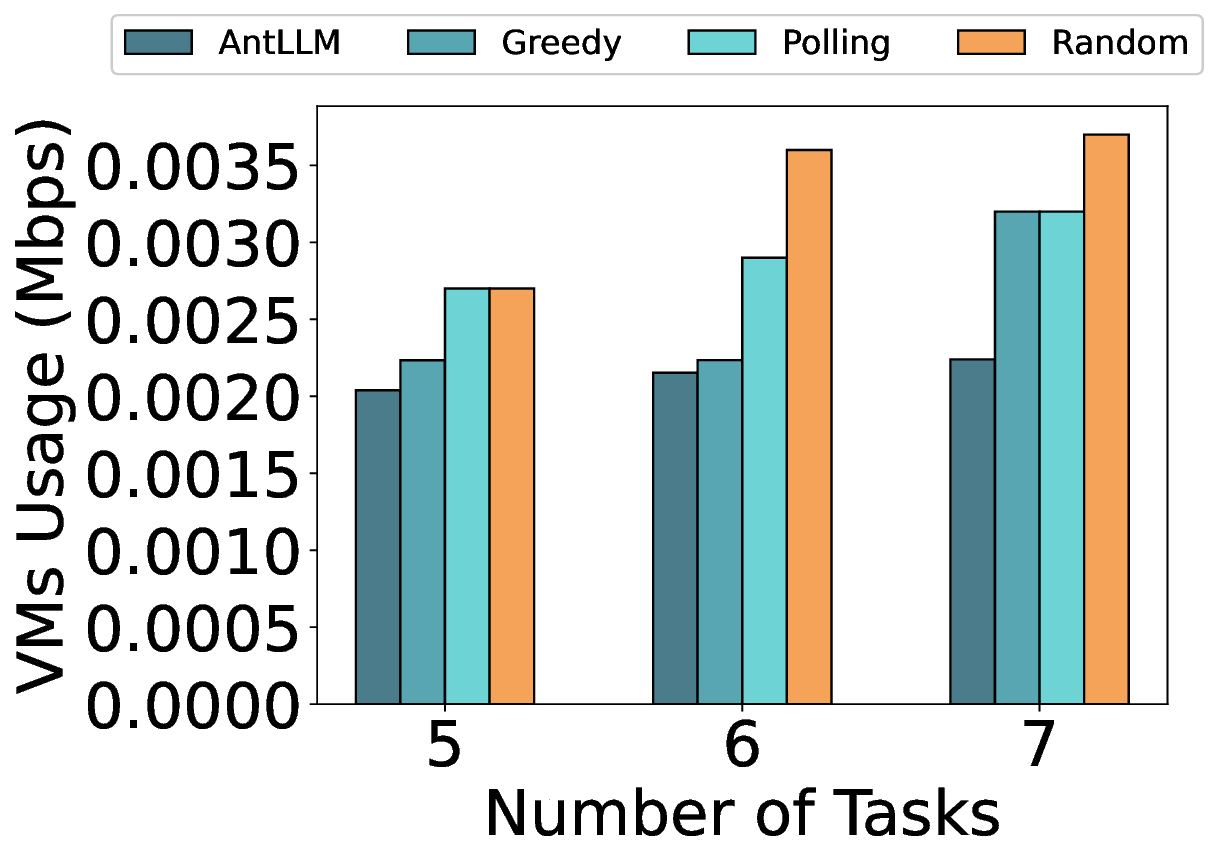} \\
        (a) CPU Usage & (b) Memory Usage & (c) Disk Usage & (d) VMs Usage
    \end{tabular}
    \caption{Cost Performance with different number of tasks}
    \label{fig:Overhead Performance with different number of tasks}
\end{figure*}

\subsection{Experimental Settings}

We evaluate our algorithms using Content-based image retrieval (CBIR) tasks. Edge servers, deployed on Huaweicloud in Beijing, Shanghai, Guiyang, and Singapore, are configured with 2 cores and 2 GB of memory to meet AI agent requirements. Beijing, Shanghai, and Guiyang servers have 40 GB of disk space and 20 Mbps peak bandwidth, while the Singapore server has 70 GB of disk space and 30 Mbps peak bandwidth. Inter-server distances are calculated using Euclidean distance.

The number of AI agents is determined by actual needs and allows flexible deployment and migration across edge servers. Balancing throughput and performance, each server can host a maximum of 10 concurrent AI agents. Each agent can access models like ``gpt-4o", ``qwen-max", ``claude 3-7-sonnet 20250404", ``glm-4-plus", and ``yi-lightning" to utilize any API within ``gpt-4o-mini" for task completion.

\textbf{Baselines:} We evaluate performance against these baselines.
\begin{enumerate}
    \item Greedy: It chooses the server with the highest score based on server capacity and AI agent resource requirements.
    \item Random: AI agents are randomly assigned to servers.
    \item Polling: Deploy AI agents sequentially. For each agent, select the first candidate server with sufficient resources.
\end{enumerate}

\subsection{Experimental Results}

\textbf{Performance with different number of servers.} Fig.\ref{fig:Delay Performance with different number of nodes} and Fig. \ref{fig:Overhead Performance with different number of nodes} illustrate the various latencies and resource costs when the number of edge servers increases. Initialization and processing times decrease with more servers, while migration time increases. This indicates that the accumulation of migration time caused by more options brought by the increase in the number of servers, and the total time is greatly affected by the cumulative migration time.

Moreover, Fig.\ref{fig:Delay Performance with different number of nodes} and Fig. \ref{fig:Overhead Performance with different number of nodes} show a trend of increasing with the increase in the number of servers. As the number of servers increases, CPU, memory, and disk usage per node decrease, while communication resource demands rise. This suggests that adding more servers reduces individual node resource consumption but increases inter-server communication needs. The performance hierarchy among different algorithms is: AntLLM $<$ Greedy $<$ Polling $<$ Random. Despite minor differences among the baselines, the AntLLM algorithm reduces the total delay by an average of 10.31\% and the resource consumption by 38.56\% compared to the baseline algorithms.

\textbf{Performance with different number of tasks.} Fig. \ref{fig:Delay Performance with different number of tasks} and Fig. \ref{fig:Overhead Performance with different number of tasks} illustrate the various latencies and resource costs when the number of tasks increases. As the number of tasks increases, both latency and resource costs rise. There are differences among each baseline, but the performance of the AntLLM algorithm is always optimal. Overall, the structural hierarchy is: AntLLM $<$ Greedy $<$ Polling $<$ Random. Compared with the baseline algorithm, the AntLLM algorithm reduces the total delay by an average of 10.64\% and the resource consumption by 49.61\%.


\section{Conclusion}

This paper presented a distributed AI agent system capable of independently deploying and migrating agents to fulfill user tasks based on user requirements. To minimize deployment and migration overhead, we implemented AntLLM, a hybrid algorithm combining ant colony optimization with LLM assistance. The experimental results showed that the AntLLM algorithm is effective. In future work, we will further consider the online incremental deployment mechanism to achieve real-time policy updates when task requirements change.

\bibliographystyle{IEEEtran}
\bibliography{references}

\end{document}